\documentclass[letterpaper, 10 pt, conference]{ieeeconf}  % Comment this line out if you need a4paper
\usepackage{amsmath,amsfonts}
\usepackage{mathrsfs}
\usepackage{bbding}
\usepackage[linesnumbered,ruled,vlined]{algorithm2e}
\SetKwInput{KwInput}{Input}
\SetKwInput{KwOutput}{Output}
\newtheorem{myDef}{Definition}

\usepackage{array}
\usepackage[caption=false,font=normalsize,labelfont=sf,textfont=sf]{subfig}
\usepackage{textcomp}
\usepackage{stfloats}
\usepackage{url}
\usepackage{verbatim}
\usepackage{graphicx}
\usepackage{cite}
\usepackage{graphics} % for pdf, bitmapped graphics files
\usepackage[caption=false]{subfig}
\usepackage{epsfig} % for postscript graphics files
\usepackage{mathptmx} % assumes new font selection scheme installed
\usepackage{times} % assumes new font selection scheme installed
\usepackage{amssymb}  % assumes amsmath package installed
\usepackage{multicol}
\usepackage{rotating}
\usepackage{color,xcolor}
\usepackage{colortbl}
\usepackage{booktabs}
\definecolor{mygray}{gray}{.85}
\usepackage{bm}
\usepackage{epstopdf}
\usepackage{multirow}
\usepackage{verbatim}
\usepackage{ulem}
\newcommand{\M}[1]{\mathbf{#1}}

\usepackage{float}
\usepackage{longtable}
\usepackage[mathscr]{euscript}

\IEEEoverridecommandlockouts                              % This command is only needed if 
\title{\LARGE \bf Graph Matching Optimization Network for Point Cloud Registration 
}
\author{Qianliang Wu, Yaqi Shen, Haobo Jiang, Guofeng Mei, Yaqing Ding, Lei Luo, Jin Xie, Jian Yang% <-this 
	\thanks{Qianliang Wu, Yaqi Shen, Haobo Jiang, Yaqing Ding, Lei Luo, Jin Xie, Jian Yang are with PCA Lab, Key Lab of Intelligent Perception and Systems for High-Dimensional Information of	Ministry of Education, and Jiangsu Key Lab of Image and Video Understanding for Social Security, School of Computer Science and Engineering, Nanjing University of Science and Technology.}
    \thanks{Guofeng Mei is with the Faculty of Engineering and Information Technology, University of Technology Sydney, Sydney NSW 2007, Australia.}
	\thanks{E-mail:\tt\small\{wuqianliang,syq,jiang.hao.bo,dingyaqing ,csjxie,csjyang\}@njust.edu.cn,\tt\small{luoleipitt@gmail
    .com,guofeng.mei@student.uts.edu.au}}
	\thanks{Corresponding authors: Jin Xie, Jian Yang}
}

\begin{document}

    \maketitle
\thispagestyle{empty}
\pagestyle{empty}

%%%%%%%%%%%%%%%%%%%%%%%%%%%%%%%%%%%%%%%%%%%%%%%%%%%%%%%%%%%%%%%%%%%%%%%%%%%%%%%%
\begin{abstract}
Point Cloud Registration is a fundamental and challenging problem in 3D computer vision. Recent works often utilize the geometric structure information in point feature embedding or outlier rejection for registration while neglecting to consider explicitly isometry-preserving constraint ($e.g.,$ point pair linked edge's length preserving after transformation) in training. We claim that the explicit isometry-preserving constraint is also important for improving feature representation abilities in the feature training stage. To this end, we propose a \underline{G}raph \underline{M}atching \underline{O}ptimization based \underline{Net}work (GMONet for short), which utilizes the graph-matching optimizer to explicitly exert the isometry preserving constraints in the point feature training to improve the point feature representation. Specifically, we exploit a partial graph-matching optimizer to optimize the super point ($i.e.,$ down-sampled key points) features and a full graph-matching optimizer to optimize fine-level point features in the overlap region.
Meanwhile, we leverage the inexact proximal point method and the mini-batch sampling technique to accelerate these two graph-matching optimizers. Given high discriminative point features in the evaluation stage, we utilize the RANSAC approach to estimate the transformation between the scanned pairs. The proposed method has been evaluated on the 3DMatch/3DLoMatch benchmarks and the KITTI benchmark. The experimental results show that our method performs competitively compared to state-of-the-art baselines.
\end{abstract}

%%%%%%%%%%%%%%%%%%%%%%%%%%%%%%%%%%%%%%%%%%%%%%%%%%%%%%%%%%%%%%%%%%%%%%%%%%%%%%%%
\section{Introduction}
Point Cloud Registration is a fundamental problem in numerous computer vision applications, such as 3D reconstruction~\cite{schonberger2016structure,choi2015robust,zhang2015visual}, localization\cite{drost2010model,zeng2017multi,yang2020teaser}, pose estimation~\cite{li2019net,zhang2014loam}, etc. Point cloud registration aims to estimate the rigid transformation between two scans. However, the partial overlapping point cloud registration is still a challenging problem due to viewpoint change or occlusion in real-world sensor data.

Recent popular deep point cloud registration methods focus on improving the registration pipeline's two key stages ($i.e.,$ point feature learning and outlier rejection). The outlier-rejection-based methods~\cite{lee2021deep,bai2021pointdsc,mei2021cotreg,shen2022reliable,jiang2021sampling} depend on the candidate correspondences computed by the point features extracted from the pre-train models ($e.g.,$ FCGF\cite{choy2019fully}). These candidate correspondences may lose the most wanted correspondences after the selection operation ($e.g.,$ KNN searching in feature space) if the point feature lack some important information. On the other hand, the feature-matching-based methods~\cite{bai2020d3feat,choy2019fully,thomas2019kpconv,huang2021predator,yu2021cofinet,qin2022geometric,li2022lepard} mainly emphasize learning more discriminative point features. Some works have tried to enhance the local features by adding translation-invariant\cite{bai2020d3feat,huang2021predator} and rotation-invariant embeddings\cite{wang2021you,li2022lepard,qin2022geometric}. To capture the geometric structure information, \cite{huang2021predator,yu2021cofinet,qin2022geometric,yew2022regtr} leverage the Transformer\cite{vaswani2017attention} to extract geometric features. However, these methods employ implicit geometric feature embedding, which lacks explicit isometric preserving constraint (i.e., edge-to-edge isometric mapping or spatial consistency) during training. We advocate that the explicit isometric preserving constraint is important for strengthening the point feature's ability to detect the overlap region. To this end, we employ two graph-matching optimizers to enhance the two-level points features in the training stage. 

%Specifically,\cite{qin2022geometric,chen2022utopic} embeds the pair-wise distance and local triplet angles into the self-attention operator to obtain rotation-invariant features. Moreover, \cite{yu2021cofinet,qin2022geometric} adopt a point-to-patch grouping strategy and utilize the optimal transport to correct the point correspondences in local patch pairs.

% - \textbf{G}raph-\textbf{M}atching-\textbf{L}earning-\textbf{Net}work 
Inspired by~\cite{bai2021pointdsc,mei2021cotreg,zhou2015factorized,gao2021deep}, we propose \textbf{G}raph \textbf{M}atching \textbf{O}ptimization based \textbf{Net}work (\textbf{GMONet} for short) to explicitly incorporate the graph matching constraint to learn the "rigid" geometric features. We utilize KPConv~\cite{thomas2019kpconv} as our feature backbone network and deploy graph-matching optimizers to enhance the isometry-preserving feature representation in training. At the coarse level, we downsample the points to super points (key points) and utilize the geometric attention layer~\cite{qin2022geometric,chen2022utopic} to generate super point features. Then, we deploy a partial graph matching optimizer to help the super points learn better to detect the overlap region. After that, we use skip links and unary layers to recover the fine-level points and features from super points features. Next, we employ another graph-matching optimizer for the points in the overlap region to help to refine point features that can help find the "rigid" correspondences. Since the cost matrices of these two-level graph-matching optimizers are built in a global scope, they can make the point feature to learn long-distance spatial consistency. We advocate that if the point features have learned enough geometric preserving information, the solution of the graph-matching optimizer could be consistent with the ground truth correspondences.

We apply two techniques to accelerate these two graph-matching optimizers. To solve the partial graph matching optimization, we transform it into an $\epsilon$-convex optimal transport problem by using the proximal point method\cite{peyre2016gromov,xu2019gromov,liu2020partial} and utilize the inexact proximal point method\cite{xie2020fast} to improve efficiency while keeping convergence. On the other hand, for the full graph matching optimizer in the overlap region, we apply the mini-batch optimal transport\cite{nguyen2022improving} strategy to accelerate the computing speed.

%The main contributions of this paper are:
Our main contributions can be summarized as follows:
\begin{itemize}
    \item We deploy two graph-matching optimizers to improve the point feature about learning isometry-preserving feature representation.
    \item We exploit the inexact proximal point method to accelerate the partial graph matching optimizer while guaranteeing convergence.
    \item We use the mini-batch technique to accelerate the graph-matching optimizer for the point feature learning in the overlap regions.
    %\item As a supplement, we propose a simple, novel, and effective interest point sampling strategy which significantly improves performance.
\end{itemize}

\section{Related work}
%In this section, first, we briefly review the deep learning and traditional methods; Second, We also summarize the recent literature on graph matching(including optimal transport problem).
% \vspace{-0.1cm}
\subsection{Traditional Methods}
ICP\cite{besl1992method} is a classical local registration method that iteratively computes the point correspondences and optimizes the least-square problem of transformation. The drawback of ICP is that it needs a good initialization to prevent the locally optimal estimation. To optimize globally, GO-ICP\cite{yang2013go} utilizes branch-and-bound optimization. However, this method is sensitive to outliers. By introducing a robust estimator ($e.g.,$ Geman-McClure), FGR\cite{zhou2016fast} improves the robustness against the outliers. Also, Teaser\cite{yang2020teaser} leverages another robust estimator ($i.e.,$ Truncated Least Squares (TLS)) and max clique to filter the inlier correspondences.
% \vspace{-0.1cm}
\subsection{Feature-Matching-Based Methods}

The famous 3DMatch\cite{zeng20173dmatch} first extracts multi-view local patch and their voxel grid of Truncated Distance Function (TDF) values, then learns 3D feature descriptors in a metric learning way. PPFNet\cite{deng2018ppfnet} uses Point Pair Features to encode local patches as inputs of the PointNet network and learns the point features by N-tuple loss. FCGF\cite{choy2019fully} designs a fully-convolutional network for computing geometric features in a single pass, which achieves a faster accurate feature extraction speed. The unsupervised PPF-FoldNet\cite{deng2018ppf} and CapsuleNet\cite{zhao20193d} exploit an encoder-decoder network to learn local feature descriptors based on point cloud reconstruction. D3feat\cite{bai2020d3feat} and Predator\cite{huang2021predator} utilize a KPConv\cite{thomas2019kpconv} module to learn translation-invariant point features. Based on Predator, CoFiNet\cite{yu2021cofinet} uses Sinkhorn's algorithm\cite{cuturi2013sinkhorn} to solve the optimal transport problem to get an optimal solution based on the initial correspondence proposal. Furthermore, GeoTransformer\cite{qin2022geometric} proposes to add edge distance and local triplet angle into the self-attention layer to learn rotation-invariant embeddings. The RGM\cite{fu2021robust} utilizes a graph feature extractor network to compute the soft correspondence matrix and convert it to a hard correspondence matrix by using a LAP solver. However, the LAP solver based on the Hungarian algorithm\cite{kuhn1955hungarian} prevents it from applying to large-scale point cloud problems.  
% \vspace{-0.1cm}
% \subsection{Learning-based Outlier Rejection Methods}

% Another class of deep learning methods focuses on the outlier rejection stage after getting the initial correspondences built from the point feature backbone. DGR\cite{choy2020deep} and 3DRegnet\cite{pais20203dregnet} use  a convolutional network and differentiable Weighted Procrustes solver to classify the inlier/outlier correspondences. \cite{shen2022reliable} generates a candidate matching map with the Dynamic Graph CNN (DGCNN)\cite{wang2019dynamic} and refines it with neighborhood consensus to predict pseudo-matching points. CEMNet\cite{jiang2021sampling} utilizes the differentiable Cross-Entropy Method\cite{amos2020differentiable} to filter reliable inlier correspondences.
% PointDSC\cite{bai2021pointdsc}  treats correspondence as points, uses max clique to get the potential high confidence inlier correspondence as key points, and conducts a one-shot max consensus to get the most confident correspondences. DHVR\cite{qi2019deep} uses the Hough Voting to discover the best transformation in  6D sparse Hough space of transformation parameters. The other concurrent work COTReg\cite{mei2021cotreg} which is an outlier rejection method utilizes an unbalanced optimal transport layer to compute reliable correspondences for point cloud registration. 
% % \vspace{-0.1cm}
\subsection{Graph Matching Methods}
Since the graph matching problem \cite{zhou2015factorized,wang2021neural} can model point-wise, pair-wise, and even more, higher order\cite{zass2008probabilistic} similarities between point sets, more and more researchers\cite{vayer2018fused,titouan2019optimal,grave2019unsupervised,alvarez2019towards,shen2021accurate,eisenberger2020deep,mandad2017variance} consider this method in image matching or network alignment problems. The classical solver for graph matching is Frank-Wolfe’s algorithm~\cite{frank1956algorithm} under the Convex-Concave Relaxation\cite{zaslavskiy2008path}. The other way is to add an entropic regularized item to the objective function to convert it to an $\epsilon$-convex problem which can be easily solved by the project gradient descent\cite{peyre2016gromov,xu2019gromov}. In order to use graph matching (or optimal transport) in large-scale problems, researchers propose the mini-batch OT (Optimal  Transport)\cite{fatras2019learning}, mini-batch UOT (Unbalanced Optimal Transport)\cite{fatras2021unbalanced}, and mini-batch POT (Partial Optimal Transport)\cite{nguyen2022improving} methods to improve efficiency while guaranteeing accuracy.

\section{Method}
% \vspace{-0.1cm}
\subsection{Problem Formulation}
Given two unordered point clouds $\M P = \{\M p_i \in \mathbb{R}^3|i=1...N\}$ and $\M Q = \{\M q_i \in \mathbb{R}^3|i=1...M\}$, where $N$ and $M$ are the different numbers of points (suppose $M>N$), the goal of the point cloud registration is to recover the rigid transformation $\M T$ consisting of $\M R \in SO(3)$ and $\M t \in \mathbb{R}^3$ that aligns $\M P$ to $\M Q$. We focus on the partial overlap point cloud registration problem~\cite{huang2021predator}. In this case, after applying the ground-truth transformation $\M T$, the overlap ratio of aligned $\M P$ and $\M Q$ is above a certain threshold $\tau$:
\begin{equation}
\small
\left| \left \{ {\M p_i}|{\M p_i} \in {\M P} \wedge  \lVert \M T(\M p_i) - {\M {NN}}(\M T(\M p_i),\M Q)\rVert_2 \leq v  \right \}\right|/|\M P|  > \tau,
%/\left|\M P\right| %% \M p_i \in \M P \land 
\end{equation}
where $\left|.\right|$ is the set cardinality, $\lVert.\rVert_2$ is the  Euclidean norm, ${\M {NN}}$ is the nearest-neighbor operator, and $v$ is a radius that depends on the point density. The overlap ratio $\tau$ is typically greater than $30\%$ in 3DMatch\cite{zeng20173dmatch} and greater than $10\%$ for low-overlap 3DLoMatch\cite{huang2021predator}. %Our paper mainly focuses on how to get more expressive point features under partial registration case.
\begin{figure*}[htbp]
      \centering
      \includegraphics[width=\textwidth]{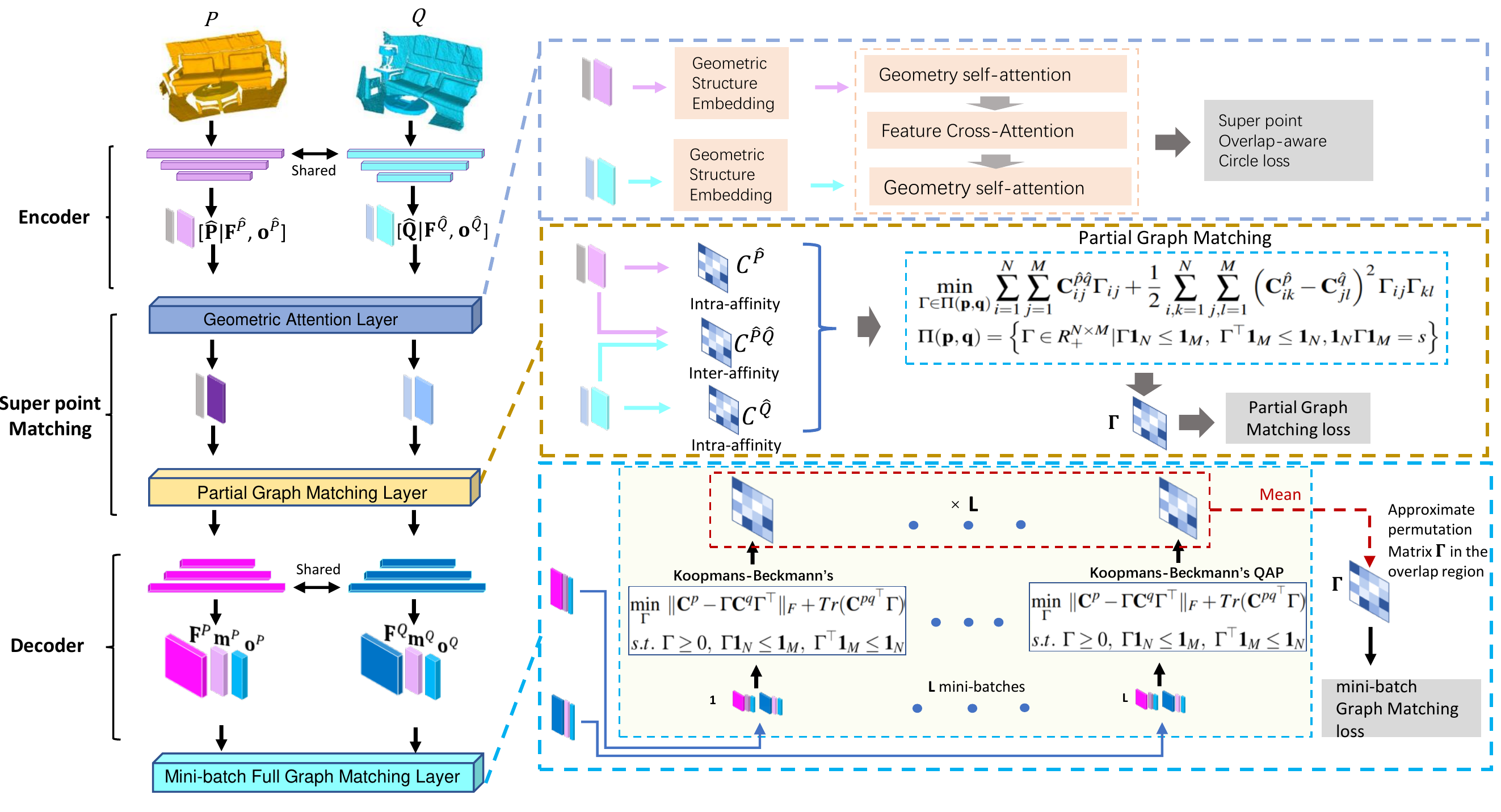}
      \setlength{\abovecaptionskip}{-0.6cm} %调整图片标题与图距离
      \caption{Overview of our proposed GMONet. First, the point clouds $\M P$ and $\M Q$ are fed to the down-sampling encoder and geometric attention layers to obtain the super points ($\hat{\M P}$,$\hat{\M Q}$), their features ($\M F^{\hat{P}}$,$\M F^{\hat{Q}}$), and overlapping scores ($\M O^{\hat{P}}$, $\M O^{\hat{Q}}$). Then, we apply the partial graph matching optimizations on super points to improve the overlap region detecting ability. Next, we use three upsampling layers to recover the fine-level points, their features ($\M F^{P}$,$\M F^{Q}$), and overlapping scores ($\M O^{P}$, $\M O^{Q}$). Lastly, a mini-batch graph matching optimizer is applied on the fine-level points in the overlap region to enhance the features' abilities for global "rigid" matching. }
      \label{framework}
      % \vspace{-0.5cm}
\end{figure*}
% \vspace{-0.1cm}
\subsection{Method overview}
The structure of our framework is illustrated in Fig.\ref{framework}. We choose KPConv\cite{thomas2019kpconv} as our feature backbone. Firstly, one point feature encoder consisting of three subsampling layers is used to downsample the given point cloud pairs to the sparse super points ($i.e.,$ $\hat{\M P} \in \mathbb{R}^{N'\times3}$ and $\hat{\M Q} \in \mathbb{R}^{M'\times3}$) and to extract associated features. Then we utilize the geometric attention layer (see section \ref{GALayer}) to give the super points embedding ($i.e.,$ $\M F^{\hat{P}} \in \mathbb{R}^{{N'}\times b}$ and $\M F^{\hat{Q}} \in \mathbb{R}^{{M'}\times b}$) and compute linear projected overlapping scores ($i.e.,$ $\M O^{\hat{P}}$ and $\M O^{\hat{Q}}$). After that, we deploy a partial graph matching optimization to enhance the super points’ ability to detect overlap regions (see section \ref{sec:PGM}). Next, we use a decoder that consists of three upsampling layers to decode the fine-level points, their corresponding features ($i.e.,$ $\M F^P \in \mathbb{R}^{N\times32}$ and $\M F^Q \in \mathbb{R}^{N\times32}$), and overlapping scores ($i.e.,$ $\M O^P$ and $\M O^Q$). Lastly, we take the mini-batch sampling to get several subsets from the overlap region and use full graph matching in each subset to refine the feature for global scaling matching (see section \ref{sec:mbGM}).
%Since the KPConv\cite{thomas2019kpconv} backbone and geometric attention layer\cite{qin2022geometric,chen2022utopic} are commonly used in the state-of-the-art methods, we do not retell them in our paper and primarily describe our two main contributed modules in the following sections.

\subsection{Point Feature Encoder}\label{sec:PEE}
Firstly, we downsample the raw point clouds to super points $\hat{\M P} \in \mathbb{R}^{N'\times3}$ and $\hat{\M Q} \in \mathbb{R}^{M'\times3}$ and generate associated features ${{\M F}^{\hat{P}}} \in R^{N'\times b}$ and ${{\M F^{\hat{Q}}} \in R^{M'\times b}}$. For super point embedding, we utilize the geometric attention layer\cite{qin2022geometric,chen2022utopic}. It encodes the local geometric structure embedding consisting of pair-wise distance (i.e., edge) and local triplet-wise angle in self-attention. It also uses cross-attention to do inter-point-cloud information exchange for overlap detection.

\underline{Local geometric structure embedding}:\label{GALayer}
For two points $p_i$ and $p_j$ in $\hat{\M P}$, the point-wise distance is $\delta_{ij} = \lVert p_i-p_j \rVert_2$. To embed the angle, we select the k nearest neighbors $\M \kappa_i$ for $p_i$. For each $\bar{p}_x \in \kappa_i$, the triplet-wise angle is computed as $\rho^x_{i,j} = \angle(\Delta_{x,i},\Delta_{j,i})$, where $\Delta_{j,i} := {\M p_j}-{\M p_i}$. We define geometric structure embedding as the combination of point-wise distance embedding and triplet-wise angle embedding:
\begin{eqnarray}
\M r_{i,j} = \M r_{i,j}^D\M W^D + max_x(\M r_{i,j,x}^A \M W^A),
\end{eqnarray}
where $\M r_{i,j}^D$ and $\M r_{i,j,x}^A$ are computed with a sinusoidal function on $\delta_{ij}/\sigma_d$ and $\rho^x_{i,j}/\sigma_a$. $\sigma_d$ and $\sigma_a$ are parameters to control the sensitivity to variations of distance and angle. $\M W^D,\M W^A \in \M R^{b\times b}$ are two linear projection layers. For points in $\hat{\M Q}$, the embeddings are computed in the same way.

\underline{Self-attention and Cross-attention}:
Given the super points' features and geometric structure embedding, we define the following geometric-aware self-attention:
\begin{eqnarray}
%&e_{i,j} = \frac{(\M x_i\M W^q)(\M x_j \M W^k + \M r_{i,j} \M W^g)^\top}{\sqrt{b}},\\
&a_{i,j} = softmax\left(\left[\frac{(\M x_i\M W^q)(\M x_j \M W^k + \M r_{i,j} \M W^g)^\top}{\sqrt{b}}\right]_{i,j}\right),\\
&\M z_i = \sum_{j=1}^{\left|\hat{\M P}\right|}a_{i,j}(\M x_j \M W^v)
\end{eqnarray}
where $W^q,W^k,W^v,W^g, \in \M R_{b\times b}$ are linear projections of queries, keys, values, and geometric structure embeddings. Given the self-attention feature $Z^{\hat{\M P}}$ and $Z^{\hat{\M Q}}$, the cross-attention layer is define as:
\begin{eqnarray}
&a_{i,j}=softmax\left(\left[\frac{(\M z_i^{\hat{\M P}} \M W^q)(\M z_j^{\hat{\M Q}} \M W^k)^\top}{\sqrt{b}}\right]_{i,j}\right),\\
&\M z_i^{\hat{\M P}} = \sum_{j=1}^{\left|\hat{\M Q}\right|}a_{i,j}(\M z_j^{\hat{\M Q}} \M W^v).
\end{eqnarray}
where $W^q,W^k,W^v \in \M R_{b\times b}$ are linear projections of queries, keys, values.

 %Additionally, one linear projection on the latent superpoint features gives overlap scores $O_{\hat{P}}$ and $O_{\hat{Q}}$. 
By three interleaved attention layers of the configuration 'self/cross/self', we get latent super point features $\M F^{\hat{P}} \in R^{{N'}\times b}$ and $\M F^{\hat{Q}} \in R^{{M'}\times b}$. To avoid symbol abuse, the initial input and final output features of 
 the attention layers are all noted as $\M F^{\hat{P}} \in R^{{N'}\times b}$ and $\M F^{\hat{Q}} \in R^{{M'}\times b}$. 

% %From super point latent features, we can compute \textit{soft correspondence} of a super point $\hat{p}_i$ (i.e., the probability that its matching point in $\hat{Q}$ lies in the overlap region):
% \begin{equation}
%      {\tilde{o}_i}^{\hat{P}} = {\M w_i^\top} {\M o^{\hat{Q}}},\ w_{ij} = softmax\left({\frac{1}{t}} \left \langle {\M f_i^{\hat{P}}},{\M f_j^{\hat{Q}}} \right \rangle \right)
% \end{equation}

\subsection{Partial Graph Matching Optimizer}\label{sec:PGM}
%Although we get the initial local geometry feature and matching score by the previous improved KPConv module,
To enhance the super points' abilities to capture isometry-preserving transformation properties, we deploy a Partial Graph Matching Optimizer (PGMO for short) to optimize the super point features. We utilize the super points ($i.e.,$ $\hat{\M P}$ and $\hat{\M Q}$) and their features ($i.e.,$ $\M F^{\hat{P}}$ and $\M F^{\hat{Q}}$) to compute the affinity matrices:
\begin{equation}
\small
\begin{split}
&{[\M C^{\hat{P}}]}_{ij}=\lVert \M f^{\hat{P}}_i-\M f^{\hat{P}}_j \rVert_2 + o^{\hat{p}_i}o^{\hat{p}_j}\alpha\lVert {\hat{\M  p}}_i-{\hat{\M p}}_j \rVert_2,\\
&{[\M C^{\hat{Q}}]}_{ij}=\lVert \M f^{\hat{Q}}_i-\M f^{\hat{Q}}_j \rVert_2 + o^{\hat{q}_i}o^{\hat{q}_j}\alpha\lVert {\hat{\M q}}_i-{\hat{\M q}}_j \rVert_2,\\
&{[\M C^{{\hat{P}\hat{Q}}}]}_{ij}=\lVert \M f^{\hat{P}}_i-\M f^{\hat{Q}}_j \rVert_2, %+ \alpha\lVert {\hat{\M p}}_i- {\hat{\M q}}_j \rVert_2*(o^{\hat{p}_i}o^{\hat{q}_j}),
\end{split}
\end{equation}
where $o^{\hat{p}_i}$ and $o^{\hat{q}_j}$ are overlapping scores of super points $\hat{\M p}_i$ and $\hat{\M q}_j$, and $\alpha$ is the hyper-parameter that controls the geometric similarity. Then we can solve the partial matching optimization to obtain the matching matrix:
\begin{equation}
\begin{split}
&\min\limits_{\Gamma \in \Pi(\M p,\M q)} \sum_{i=1}^N\sum_{j=1}^M \M C^{\hat{P}\hat{Q}}_{ij}\Gamma_{ij} + \frac{1}{2} \sum_{i,k=1}^N \sum_{j,l=1}^M \left({\M C_{ik}^{\hat{P}}}-{\M C_{jl}^{\hat{Q}}}\right)^2 \Gamma_{ij}\Gamma_{kl} \\
&=\min\limits_{\Gamma \in \Pi(\M p,\M q)} \langle \M C^{\hat{P}\hat{Q}},\Gamma \rangle + \langle \M L(\M C^{\hat{P}},\M C^{\hat{Q}}, \Gamma), \Gamma \rangle,\\
%&\Pi(\M p,\M q) = \left \{ \Gamma \in R_+^{N \times M}| \Gamma \M 1_N \leq \M 1_M,\  \Gamma^\top \M 1_M \leq \M 1_N, \M 1_N\Gamma\M 1_M=s \right \}
\end{split}\label{pgmp2}
\end{equation}
where $\langle.\rangle$ is inner product, $\M L(\M C^{\hat{P}},\M C^{\hat{Q}}, \Gamma) = \left[{\M L_{jj'}}\right] \in \mathbb{R}^{N{\times}M}$, each ${\M L_{jj'}}={\sum_{i,{i'}}}\mathcal{L}\left(\M C^{\hat{P}}_{ij},\M C^{\hat{Q}}_{{i'}{j'}}\right){\Gamma_{ii'}}$, and $\mathcal{L}(a,b) =\frac{1}{2}(a-b)^2$.
The admissible couplings $\Pi(\M p,\M q)$ are defined as $\{ \Gamma \in R_+^{N \times M}| \Gamma \M 1_N \leq \M 1_M,\  \Gamma^\top \M 1_M \leq \M 1_N, \M 1_N\Gamma\M 1_M=s   \}$. 
%The affinity matrices $\M C^{\hat{P}}$, $\M C^{\hat{Q}}$, and $\M C^{\hat{P}\hat{Q}}$ are computed from super points' coordinates and features:
 The empirical distributions $(\M p,\M q) \in \Sigma_N\times\Sigma_M$, ${\Sigma_N}$ is a histogram of $N$ bins with $\left \{\M p \in R^N_+,\sum_i p_i =1\right \}$. We utilize the uniform distribution to initialize $(\M p,\M q)$. The partial transport mass $s$ is computed from the super points’ anchored patch\cite{qin2022geometric} pairs whose overlap ratio is higher than 10$\%$.

% Since the proximal point method is an efficient algorithm to solve the non-convex problem, motivated by~\cite{xie2020fast}, in the n-th iteration, we update correspondence matrix $\Gamma$ as: 
% \begin{equation}
% \small
% \Gamma^{n+1} = \arg\min\limits_{\Gamma \in \Pi(\M p,\M q)} \langle \M C^{\hat{P}\hat{Q}},\Gamma \rangle + \langle \M L(\M C^{\hat{P}},\M C^{\hat{Q}}, \Gamma), \Gamma \rangle + \epsilon KL(\Gamma||{\Gamma^n}),\label{ppa}
% \end{equation}
% where $KL(a||b)=alog(\frac{a}{b}) -a +b$.

Inspired by~\cite{peyre2016gromov}, by using mirror descent and Bregman projection ($i.e.,$ both the gradient and the projection are computed in the $KL$ metric) and setting the learning rate to $\frac{1}{\epsilon}$, we can transform problem~(\ref{pgmp2}) to a new $\epsilon$-convex entropic regularized optimal transport problem: 
\begin{equation}
\Gamma^{n+1} = \arg\min\limits_{\Gamma \in \Pi(\M p,\M q)} \langle \bar{\M C^n} - \epsilon log \Gamma^n ,\Gamma \rangle+ \epsilon H(\Gamma),\label{convexot}
\end{equation}
where $\bar{\M C^n}=L(\M C^{\hat{P}},\M C^{\hat{Q}}, \Gamma^n)+\M C^{\hat{P}\hat{Q}}$ and the entropy $H(\Gamma)=-\sum_{i,j=1}^N \Gamma_{i,j}(log(\Gamma_{i,j})-1)$. According to Proposition 5 in~\cite{benamou2015iterative}, problem~(\ref{convexot}) is a partial transport problem with inequality constraints, which needs to use the Dykstra’s algorithm~\cite{dykstra1983algorithm} to solve it. To accelerate the computing speed, motivated by the inexact proximal point algorithm (IPOT) in~\cite{xie2020fast}, the inner number of Dykstra's iterations L is set to 1.
% \vspace{-0.2cm}
% \begin{algorithm}[htbp]
% \caption{Inexact Proximal Optimal Transport for Partial Graph Matching (IPOT-PGM)}
% \small
% \DontPrintSemicolon
%   \KwInput{Intra-affinity matrix $\M C^{\hat{P}}$ and $\M C^{\hat{Q}}$, Inter-affinity matrix $\M C^{\hat{P}\hat{Q}}$, Overlap ratio $s$, Probabilities $\{\M p,\M q\}$ on support points $\{\hat{\M P}, \hat{\M Q}\}$, Entropy regularized weight $\epsilon$.}
%   \KwOutput{Correspondence matrix $\Gamma^{K}$}

%   Initialize $\Gamma^0={\M p}{\M q}^\top$\\
%   \For{$n=0:K-1$}
%   {
%    \tcc{Efficiently computing ${ \bar{\M C^n}}$ in Eqn.(\ref{convexot}), see Proposition 1 in~\cite{peyre2016gromov}. $f_1(a) = a^2, f_2(b) = b^2, h_1(a) = a, h_2(b) = 2b$}
%     $Cost^{\hat{P}\hat{Q}} = f_1(\M C^{\hat{P}}){\M p}{\M 1_M^T} + {\M 1_N}{\M q}^Tf_2({\M C^{\hat{Q}}})^T$\\
%     ${ \bar{\M C^n}} = Cost^{\hat{P}\hat{Q}} - h_1(\M C^{\hat{P}})\Gamma^nh_2(\M C^{\hat{Q}})^T + \M C^{\hat{P}\hat{Q}}$\\
%     Set $\M G = exp(\frac{-\bar{\M C^n}}{\epsilon})\odot\Gamma^n$ \\
%     \tcp{Usually set L=1 for Inexact Proximal}
%     \For{$l=1,2,3...L$}
%     {
%     \tcp{Dykstra's iterations}
%     $\M G = diag\left(min\left(\frac{\M p}{\M G\M 1_M},{\M 1_N}\right)\right)\M G$\\
%     $\M G = {\M G} diag\left(min\left(\frac{\M q}{\M G^\top\M 1_N},{\M 1_M}\right)\right)$\\
%     $\M G = {\M G}\odot{\frac{s}{\M 1_N^\top{\M G}\M 1_M}}$
%     }
%   }
%   $\Gamma^{K}=\M G$
  
% \end{algorithm}\label{IPOT-PGM}
%\vspace{-0.3cm}

\subsection{Point Feature Decoder}
Given the super point features, we need to recover the original resolution point features. We leverage the NN upsampling and skip connections from the downsampling layers to form the decoder. We firstly concatenate the super point features $F^{\hat{P}}|F^{\hat{Q}}$, and the overlap scores ${O}^{\hat{P}}|{O}^{\hat{Q}}$, then go through upsampling decoder to get the fine-level ones: $F^P|F^Q$ and $O^P|O^Q$. 

\subsection{Mini-batch Full Graph Matching Optimizer}\label{sec:mbGM}
To enhance the fine-level points’ abilities to capture isometry-preserving transformation properties in the overlapping regions, we exploit a Full Graph Matching Optimizer Optimizer (FGMO for short) to optimize the point features:
\begin{equation}
% \begin{split}
\min\limits_{\Gamma \in {\hat{\Pi}(\M p,\M q)}} {\M J}_{gm}({\Gamma}) =  \min \limits_{\Gamma}  \lVert \M C^P - \Gamma \M C^Q \Gamma^\top\rVert_F + Tr(\M C^{{PQ}^\top}\Gamma),   \\
    %&s.t.\  \Gamma \in \left\{0,1\right\}^{N\timesM},\ \Gamma \M 1_N \leq \M 1_M,\  \Gamma^\top \M 1_M \leq \M 1_N
% &s.t.\  \Gamma \ge 0,\ \Gamma {\M 1_N} \leq {\M 1_M},\  \Gamma^\top {\M 1_M} \leq {\M 1_N},
% &\hat{\Pi}(\M p,\M q) = \{ \Gamma \in R_+^{N \times M}| \Gamma \ge 0,\ \Gamma {\M 1_N} \leq {\M 1_M},\  \Gamma^\top {\M 1_M} \leq {\M 1_N} \},
% \end{split}
\label{pgmp}
\end{equation}
where $\M C^P$ and $\M C^Q$ are two affinity matrices of two graphs generated from points coordinates and features. $\M C^{PQ}$ is the inter-graph affinity matrix or cost matrix. $\hat{\Pi}(\M p,\M q)$ is defined as $\{ \Gamma \in R_+^{N \times M}| \Gamma \ge 0,\ \Gamma {\M 1_N} \leq {\M 1_M},\  \Gamma^\top {\M 1_M} \leq {\M 1_N} \}$. The definitions of $\M C^P$, $\M C^Q$, and $\M C^{PQ}$ are as follows:
\begin{equation}
\begin{split}
&{[\M C^P]}_{ij}=\lVert \M f^P_i-\M f^P_j \rVert_2 + \alpha\lVert \M p_i- \M p_j \rVert_2,\\
&{[\M C^Q]}_{ij}=\lVert \M f^Q_i-\M f^Q_j \rVert_2 + \alpha\lVert \M q_i-\M q_j \rVert_2,\\
&{[\M C^{PQ}]}_{ij}=\lVert \M f^P_i-\M f^Q_j \rVert_2,% + \alpha\lVert \M p_i-\M q_j \rVert_2,
\end{split}
\end{equation}
where $\alpha$ is a hyper-parameter that controls the geometric similarity. 

% The optimizing object in this problem is non-convex since the existence of the quadratic term (i.e., the first part of object in \eqref{pgmp}). A typical way to solve this problem is to add $M-N$ dummy nodes to point cloud $\M P$ to make a square matching matrix $\Gamma$. Then a path-following strategy~\cite{zaslavskiy2008path} can be adopted to make gradually non-convex optimization to solve this problem, and Frank-Wolfe’s algorithm~\cite{frank1956algorithm} is used 
% in this strategy. For the big graph matching problem~($e.g.,$ $M>100$), Factorized Graph Matching~(FGM)~\cite{zhou2015factorized} may be an available tool. However, the matching matrix needs to be full rank~($i.e.,$ overlap ratio is $100\%$) in the concave-convex relaxation, so we use it to optimize the full graph matching of correspondences in the overlap region. 
Considering the large scale of fine level point cloud, we choose a reduced path following algorithm to efficiently solve the problem~(\ref{pgmp}). Furthermore, we take the mini-batch method~\cite{fatras2021unbalanced,nguyen2022improving} to accelerate solving the problem~(\ref{pgmp}) in the overlap region:
\begin{myDef}

(Mini-batch Graph Matching) For subset size, $1\leq m \leq min(M,N)$ and subsets number $K \ge 1$, $\M p_1^m$,...,$\M p_K^m$ and $\M q_1^m$,...,$\M q_K^m$ are subsets  that are sampled from the overlap region of point clouds $\M P$ and $\M Q$, respectively. The mini-batch transport plan is:
% \vspace{-0.3cm}
\begin{equation}
\small
\begin{split}
&\Gamma^{m,K,s} = \frac{1}{K}\sum_{i=1}^K\Gamma(\M p_i^m, \M q_i^m),\\
&\Gamma(\M p_i^m,\M q_i^m) = \arg\min\limits_{\Gamma \in \Pi(\emph{\M p_i^m},\emph{\M q_i^m})} {\lVert \M C^{ \M p_i^m} - \Gamma \M C^{\M q_i^m}} \Gamma^\top\rVert_F + Tr(\M C^{{{\M p_i^m}{\M q_i^m}}^\top}\Gamma),\\
%\langle \M C^{{\M p_i^m}{\M q_i^m}},\Gamma \rangle + \langle \M L(\M C^{ \M p_i^m},\M C^{ \M q_i^m}, \Gamma), \Gamma \rangle,
\end{split}
\end{equation}
where ($\emph{\M p_i^m},\emph{\M q_i^m}$) are two empirical distributions computed from two subsets $\M p_i^m$ and $\M q_i^m$. 
%Note that $\Gamma(p_i^m,q_i^m)$ is expanded to size of $N\times M$ and the zero padded entries are the unsampled indices.
\end{myDef}

We uniformly sample $K$ subsets from the fine-level overlap region, and each subset contains $m$ points. The average of $K$ mini-batch solutions would give an approximation of ground truth correspondences.
% \vspace{-0.1cm}
\subsection{Loss Function}
\textbf{Coarse-level Overlap-aware Circle loss:}
Inspired by \cite{qin2022geometric}, the overlap ratio of patches anchored on the super points can weight the positive matching in loss to avoid the issue that the circle loss weights the positive samples equally. Given a pair of aligned coarse-level super points $\hat{\M P}$ and $\hat{\M Q}$. A pair of super points is positive if their anchor patch shares at least a 10\% overlap ratio and negative if there is no overlap. All other pairs are dropped. For a super point $\hat{p}_i$, which at least has one positive patch in $\hat{\M Q}$, we define the super points in $\hat{\M Q}$ within radius $r_{\hat{p}}$ as $\M \xi_{p}(\hat{p}_i)$ and the super points outside a larger radius $r_n$ as  $\M \xi_n(\hat{p}_i)$. Inspired by \cite{yu2021cofinet,qin2022geometric}, we sampled $n_p$ points from $\hat{P}_{\hat{p}}$, and the circle loss can be defined as follows:
\begin{equation}
\small
{\mathscr{L}_{coc}^{\hat{\M P}}} = \frac{1}{n_p} \sum_{i=1}^{n_p}log \left[ 1+\sum_{j\in \xi_{p}(\hat{p}_i)} e^{\lambda_i^j\gamma(d_i^j-\Delta_p)} + \sum_{k\in \xi_{n}(\hat{p}_i)} e^{\gamma(\Delta_n - d_i^k)} \right],
\end{equation}
where $d_i^j = \lVert \M f^{\hat{p}_i} - \M f^{\hat{q}_j}\rVert_2$, $\gamma$ is a hyper-parameters and $\lambda_i^j$ is the overlap ratio of patches anchored on super point $\hat{\M p}_i$ and $\hat{\M q}_j$. Two empirical margins are defined as $\Delta_p:=0.1$ and $\Delta_n:=1.4$. The reverse loss $\mathscr{L}_{coc}^{\hat{\M Q}}$ is also defined in the same way and the final circle loss is $\mathscr{L}_{coc} = \frac{1}{2} (\mathscr{L}_{coc}^{\hat{\M P}} + \mathscr{L}_{coc}^{\hat{\M Q}} )$.

\textbf{Coarse-level Partial Graph Matching loss:}
We calculate the intra-affinity and inter-affinity matrix for Eqn.\eqref{pgmp2} based on the coarse-level super points and features. The solution of the partial graph matching optimization (i.e., Eqn.\eqref{pgmp2}) can be treated as soft matching scores. The supervision on the matching score can be cast a binary classification, and we define a cross-entropy loss:
% \vspace{-0.2cm}
\begin{eqnarray}
\mathscr{L}_{cpgm}^{\hat{\M P}} = \frac{1}{|\hat{\M P}|}\sum_{i=1}^{\hat{\M P}}\bar{m}^{\hat{p}_i}log(m^{\hat{p}_i})+(1-\bar{m}^{\hat{p}_i})log(1-m^{{\hat{p}}_i}),
\end{eqnarray}
where ground truth $\bar{m}^{\bar{p}_i}$ is based on the overlap ratio of patch pairs that $\bar{m}^{\hat{p}_i}$ is 1 if the overlap ratio is greater than $10\%$, otherwise 0. The reverse loss $\mathscr{L}_{cpgm}^{\hat{\M Q}}$ is also defined in the same way, and the final loss is $\mathscr{L}_{cpgm} = \frac{1}{2} (\mathscr{L}_{cpgm}^{\hat{\M P}} + \mathscr{L}_{cpgm}^{\hat{\M Q}})$.

\textbf{Fine-Level mini-batch Graph Matching loss:}

For every sample subset in the overlap region, the loss supervising the predicted matching scores is defined as:
% \vspace{-0.2cm}
\begin{eqnarray}
\mathscr{L}_{mbm}^{\M P} = \frac{1}{|\M P|}\sum_{i=1}^{\M P}\bar{m}^{p_i}log(m^{p_i})+(1-\bar{m}^{p_i})log(1-m^{p_i}),
\end{eqnarray}
where $\bar{m}^{p_i}$ is ground truth and $m^{p_i}$ is solution of Eqn.\eqref{pgmp}. The reverse loss $\mathscr{L}_{mbm}^{\M Q}$ is also defined in the same way, and the final loss is $\mathscr{L}_{mbm} = \frac{1}{2} (\mathscr{L}_{mbm}^{\M P} + \mathscr{L}_{mbm}^{\M Q})$.

\textbf{Fine-level overlap loss:} To supervise the predicted overlap score on fine-level points, we use a binary classification loss for the overlap probability:
% \vspace{-0.2cm}
\begin{eqnarray}
\mathscr{L}_{o}^{\M P} = \frac{1}{|\M P|}\sum_{i=1}^{\M P}\hat{o}^{p_i}log(o^{p_i})+(1-\hat{o}^{p_i})log(1-o^{{p}_i}),
\end{eqnarray}
where ground truth label $\hat{o}^{p_i}$ is defined as
% \vspace{-0.2cm}
\begin{eqnarray}
    \hat{o}^{p_i} = 
    \begin{cases}
    1, \ \lVert \M T(\M p_i) - {\M {NN}}(\M T(\M p_i),\M Q)\rVert_2 < \epsilon_o\\
    0, \ otherwise
    \end{cases},
\end{eqnarray}
with overlap threshold $\epsilon_o$. The reverse loss $\mathscr{L}_{o}^{\M Q}$ is computed in the same way and  $\mathscr{L}_o = \frac{1}{2}(\mathscr{L}_{o}^{\M P} + \mathscr{L}_{o}^{\M Q})$.
% Similarily, we also use a binary classification loss to supervise the matching score:
% % \vspace{-0.2cm}
% \begin{eqnarray}
%     \mathscr{L}_{m}^{\M P} = \frac{1}{|\M P|}\sum_{i=1}^{\M P}\hat{m}^{p_i}log(m^{p_i})+(1-\hat{m}^{p_i})log(1-m^{{p}_i}),
% \end{eqnarray}
% where ground truth labels $\hat{m}^{p_i}$ are generated on the fly
% % \vspace{-0.2cm}
% \begin{eqnarray}
%     \hat{m}^{p_i} = 
%     \begin{cases}
%     1, \ \lVert \M T(\M p_i) - {\M {NN}_{F}}(\M T(\M p_i),\M Q)\rVert_2 < \epsilon_m\\
%     0, \ otherwise
%     \end{cases},
% \end{eqnarray}
% via nearest neighbour search NN$_F$(.,.) in feature space with overlap threshold $\epsilon_m$. The reverse loss $\mathscr{L}_{m}^{\M Q}$ is defined in the same way and $\mathscr{L}_m = \frac{1}{2}(\mathscr{L}_{m}^{\M P} + \mathscr{L}_{m}^{\M Q})$. 
%The circle loss $\mathscr{L}_{c}$ defined on fine level points $\M P$ and $\M Q$ is similar to $\mathscr{L}_{coc}$ except without the overlap ratio coefficient $\lambda_i^j$.
The final loss is defined as
% \vspace{-0.1cm}
\begin{eqnarray}
\mathscr{L} = \lambda_c(\mathscr{L}_{coc} + \mathscr{L}_{cpgm}) + \lambda_f(L_{mbm} + \mathscr{L}_o),
\end{eqnarray}
where $\lambda_c$ and $\lambda_f$ are the weights of the coarse-level and the fine-level losses, respectively. Following \cite{yu2021cofinet}, we set $\lambda_c = \lambda_f = 1$.
% \vspace{-0.2cm}
\section{Experiments}
\subsection{Experimental Settings}
Following \cite{huang2021predator}, we evaluate the proposed GMONet on indoor datasets 3DMatch\cite{zeng20173dmatch} and 3DLoMatch~\cite{huang2021predator} and outdoor KITTI odometry\cite{geiger2013vision} benchmark.

\textbf{Implementation and training:}
The proposed GMONet is implemented and tested with PyTorch~\cite{paszke2019pytorch} on Xeon(R) Gold 6230 and one NVIDIA RTX TITAN GPU. The network is trained 30 epochs on the 3DMatch/3DLoMatch dataset and 150 epochs on the KITTI odometry benchmark, all with Adam optimizer. The learning rates for 3DMatch/3DLoMatch and KITTI are set to 1e-4 and 5e-2, respectively. The batch size, weight decay, and momentum are set as 1, 1e-6, and 0.98, respectively. 3DMatch and KITTI's matching radii are set to 5cm and 30cm, respectively. The hyper-parameter $\alpha$ in \ref{sec:PGM} is set to 0.01.
%The other parameters are the same as \cite{huang2021predator}.

\begin{figure*}[htbp]
      \centering
      \includegraphics[width=\textwidth]{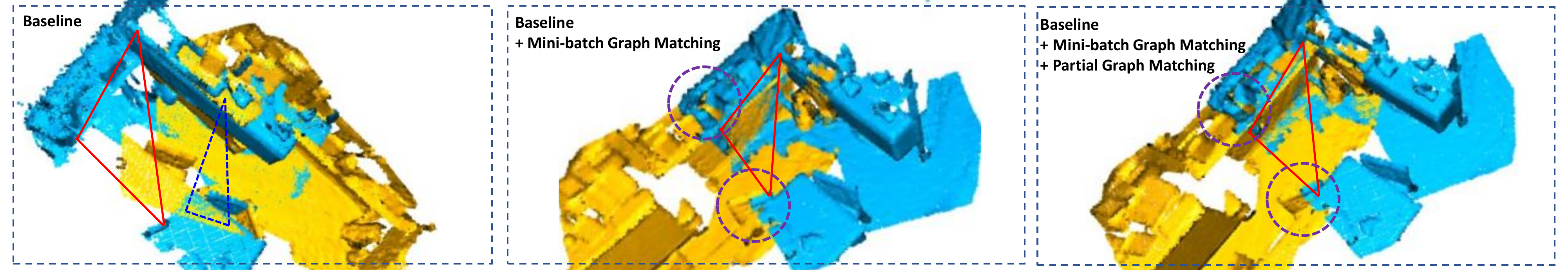}
      \caption{Visualization of the effective role of coarse-level partial graph matching constraint and fine-level graph matching constraint.}
      \label{fig:3dmatch_ablay}
      \vspace{-0.3cm}
\end{figure*}

% \vspace{-0.1cm}
\subsection{Indoor dataset: 3DMatch and 3DLoMatch}
\textbf{Dataset.} 3DMatch\cite{zeng20173dmatch} contains 62 scenes, divided into 46, 8, and 8 for training, validating, and testing, respectively. The overlap ratio of scanned pairs in 3DMatch is greater than $30\%$, while $10\%$-$30\%$ in 3DLoMatch.

\textbf{Metrics.}
Following\cite{zeng20173dmatch}, we report performance with three metrics: (1) rigid Registration Recall (RR), the fraction of point cloud pairs whose correspondence RMSE below 0.2m. (2) Relative Rotation Error (RRE), the geodesic distance between estimated and ground truth rotation matrices. (3) Relative Translation Error (RTE), the Euclidean distance between the estimated and ground truth translation. RRE and RTE are calculated on the successfully matching scan pairs. 
 
% Since the high inlier ratio does not necessarily lead to high registration recall as the correspondences could cluster together as noted in \cite{huang2021predator}, we skip the other two metrics Feature Matching Recall (FMR) and Inlier Ratio (IR) in the benchmark comparison. The RTE and RRE are defined as the mean errors of successfully registered ones.
%and only use them in ablation studies.

\textbf{Interest point sampling.}
In the evaluation stage, we multiply the matching scores (normalized inner-product matrix from point features) and overlapping scores as the inlier confidence probabilities of points. Then a random sampling based on the confidence probability is applied to obtain the candidate point correspondences. 

\textbf{Results.}
We compare GMONet to recent feature-matching-based methods (FCGF\cite{choy2019fully}, D3Feat\cite{bai2020d3feat}, Predator\cite{huang2021predator}, CoFiNet\cite{yu2021cofinet}, GeoTransformer\cite{qin2022geometric}). We adopt the same sampling strategy as GMONet to evaluate baseline GeoTransformer\cite{qin2022geometric} for a fair comparison. 

Tab.\ref{res:3DMatch} shows that GMONet achieves the best registration recall to 92.1\% on 3DMatch and 73.2\% on 3DLoMatch with a sampling of 1000 points. Our method achieves relatively lower RTE and RRE on both 3DMatch and 3DLoMatch benchmarks. This reveals that, by adding two graph-matching optimizers in the learning stage, the point features indeed learn the isometry-preserving features and help select the "rigid" candidate corresponding point more precisely.

\begin{table}
	\caption{Quantitative results on the 3DMatch and 3DLoMatch benchmarks. The best results are highlighted in bold, and the second best results are underlined.}
	\label{res:3DMatch}
	\centering
	
	\resizebox{0.95\linewidth}{!}{
	\huge
	\begin{tabular}{c|ccccc|ccccc}
		\toprule

% 		&&&3DMatch&&&&&3DLoMatch&&&
		& \multicolumn{5}{c}{3DMatch} & \multicolumn{5}{c}{3DLoMatch} \\
% 		\multicolumn{11}{c}{ &3DMatch&3DLoMatch} 
		\midrule
		\# Samples&5000&2500&1000&500&250&5000&2500&1000&500&250\\
		\midrule
		\multicolumn{11}{c}{RR ($\%$)$\uparrow$}           \\
		\midrule
		FCGF\cite{choy2019fully}&85.1&84.7&83.3&81.6&71.4&40.1&41.7&38.2&35.4& 26.8\\
		D3Feat\cite{bai2020d3feat}&81.6&84.5&83.4&82.4&77.9&37.2&42.7&46.9&43.8&39.1\\
		Predator\cite{huang2021predator}&89.0&89.9&90.6&88.5&86.6&59.8&61.2&62.4&60.8&58.1\\
		CoFiNet\cite{yu2021cofinet}&89.3&88.9&88.4&87.4&87.0&67.5&66.2&64.2&63.1&61.0\\
        GeoTransformer\cite{qin2022geometric}&\textbf{91.4}&\uline{91.3}&\uline{91.4}&\uline{90.8}&\textbf{90.4}&\textbf{72.3}&\uline{72.0}&\uline{71.7}&\uline{72.3}&\textbf{71.3}\\
		
		\midrule
		GMONet&\uline{91.3}&\textbf{91.8}&\textbf{92.1}&\textbf{91.0}&\uline{89.5}&\uline{69.0}&\textbf{72.4}&\textbf{73.2}&\textbf{72.6}&\uline{69.9}\\
        
        \midrule
        \multicolumn{11}{c}{RTE (m)$\downarrow$} \\
        \midrule
        FCGF\cite{choy2019fully}&0.066 &- &\uline{0.066} &- &- &- & -&0.105 & -&- \\
        D3Feat\cite{bai2020d3feat}&0.067 &- & -& -& &- & -& -&- &- \\
        Predator\cite{huang2021predator}&\uline{0.064} &\uline{0.063} &\uline{0.068}&\textbf{0.069}&0.076& 0.091&\uline{0.089}&\uline{0.092}& 0.102&0.108 \\
        CoFiNet\cite{yu2021cofinet}& \uline{0.064}& \uline{0.063}& 0.069& \uline{0.070}&\uline{0.074} &\uline{0.090} & 0.095& 0.096&\uline{0.099}&\uline{0.107} \\
        GeoTransformer\cite{qin2022geometric}&0.070 &0.069 &0.071 &\uline{0.070}&\textbf{0.072} &0.097 &0.099 &  0.099& 0.101&\textbf{0.101} \\
        \midrule
         GMONet& \textbf{0.061}& \textbf{0.062}& \textbf{0.063}& 0.071&0.077&\textbf{0.089}& \textbf{0.089}&\textbf{0.088}&\textbf{0.093}&0.109 \\        
        \midrule
        \multicolumn{11}{c}{RRE ($^\circ$)$\downarrow$} \\
        \midrule
        FCGF\cite{choy2019fully}&1.949 &- & \uline{2.060}& -&- &- &- &3.820&-&- \\
        D3Feat\cite{bai2020d3feat}&2.161 &- &- & -&- & -&- &- &- & -\\
        Predator\cite{huang2021predator}&\uline{1.847}&\uline{1.869}&\uline{1.998}& 2.169&\textbf{2.468} &\uline{3.156}& \uline{3.124}&3.368&3.675&3.927 \\
        CoFiNet\cite{yu2021cofinet}& 2.002& 2.124& 2.281& 2.302& 2.486& 3.271& 3.415& 3.520& 3.513&3.748 \\
        GeoTransformer\cite{qin2022geometric}& 2.021& 2.041&2.072 &\textbf{2.019} &2.134 & 3.238& 3.383& \uline{3.267}& \textbf{3.298}&\textbf{3.472} \\
        \midrule
        GMONet&\textbf{1.841} & \textbf{1.857}& \textbf{1.857}& \uline{2.059}&\uline{2.500} &\textbf{2.856} &\textbf{2.959} & \textbf{2.937}&\uline{3.300}&\uline{3.637} \\     
		\bottomrule
	\end{tabular}}
    % \vspace{-0.3cm}
\end{table}  
% \vspace{-0.2cm}
\begin{table}
	\caption{Ablation of the network architecture on 3DMatch/3DLoMatch benchmark. Tested with Samples=1000.}
	\label{res:ablay}
	\centering
	\resizebox{0.5\textwidth}{!}{
	\Huge
    \begin{tabular}{cc|ccc|ccc}
		\toprule
   
        \multicolumn{2}{l}{} &  &3DMatch &    &  &3DLoMatch&  \\
        \midrule
        FGMO&PGMO&RR (\%)&RRE ($^\circ$)&RTE (cm)&RR (\%)&RRE ($^\circ$)&RTE (cm)\\
        \midrule 
          &&90.02&2.014&0.065&68.2&3.070&0.092\\
          $\surd$&&91.40&1.875&0.063&71.2&2.860&0.090  \\
          &          $\surd$& 90.60 & 1.946&0.063& 68.8&3.062&0.089\\
          $\surd$&$\surd$&92.10&1.857&0.063&73.2&2.937&0.088 \\
        \bottomrule
    \end{tabular}
    }
    \vspace{-0.2cm}
\end{table}
\begin{table}
	\caption{Quantitative results on the KITTI odometry benchmark. The best results are highlighted in bold, and the second-best results are underlined.}
	\label{res:KITTI}
	\centering
	\resizebox{0.8\linewidth}{!}{
	\small
	\begin{tabular}{c|ccc}
		\toprule
	
		Method&RTE (cm)$\downarrow$&RRE ($^\circ$)$\downarrow$&RR ($\%$)$\uparrow$\\
		\midrule
		3DFeat-Net\cite{yew20183dfeat} &25.9 &\textbf{0.25}& 96.0\\
		FCGF\cite{choy2019fully}&9.5&0.30&\uline{96.6}\\
		D3Feat\cite{bai2020d3feat}&7.2&0.30&99.8\\
		%SpinNet&9.9&0.47&99.1\\
		Predator\cite{huang2021predator}&\textbf{6.8}&\uline{0.27}&\textbf{99.8}\\
		CoFiNet\cite{yu2021cofinet}&8.2&0.41&\textbf{99.8}\\
		GeoTransformer\cite{qin2022geometric}&7.4&0.27& \textbf{99.8}\\
		GMONet&\textbf{6.8}&\uline{0.27}&\textbf{99.8}\\
		\bottomrule
	\end{tabular}}
	\vspace{-0.6cm}
\end{table}
\textbf{Ablation studies.}
In the ablation studies, we take KPConv\cite{thomas2019kpconv} as the backbone plus geometric attention layer, coarse-level overlap-aware circle loss, and fine-level overlap loss as the baseline. Then we add two levels of graph matching optimizer to conduct extensive experiments to ablate how these two constraints improve the feature representation. The default number of sampling points is set to 1000. Tab.\ref{res:ablay} illustrates that by adding partial graph matching constraint on the coarse-level super points, the registration recall increases 0.58 percent points (pp) on 3DMatch and 0.6 pp on 3DLoMatch. The mini-batch graph matching constraint on fine-level points improves by 1.38 pp on 3DMatch and 3.0 pp on 3DLoMatch, respectively. These two constraints improve 2.08 pp and 5.0 pp on 3DMatch and 3DLoMatch, respectively.

As visualized test case in Fig.\ref{fig:3dmatch_ablay}, without explicit isometric preserving constraints, based on the point features, RANSAC estimation would prefer such correspondences, $e.g.,$ edges near the vertexes of the \textcolor{blue}{deep blue triangle}, that gives the max number of "looks likely" correspondences (see the left column) in the overlap region. However, by explicitly adding two isometric preserving constraints, the point feature would recognize the critical correspondences ($e.g.,$ correspondences near the three vertexes of the \textcolor{red}{red triangle} in the middle column) even though the candidate corresponding points are far from each other in the euclidean space. Moreover, the graph matching optimization on coarse-level points further improves the registration's precision (see points in the \textcolor{purple}{purple circle} in middle and right columns). This illustrates that the two levels of graph matching optimizers help strengthen the points' abilities to maintain isometry-preserving in feature space.

\textbf{Computational Complexity.}
The running time of the two optimizers is listed in Tab.\ref{res:comlexsity}. Since we deploy two proposed optimizers only in the training stage and turn them off when inference, the run time is not a burden. The running time of the Partial Graph Matching Optimizer depends on the number of down-sampled super points, while Mini-batch Full Graph Matching Optimizer depends on the size of each sampled subset. Our default sampling number for each subset is set to 128.

\begin{table}
	\caption{Runtime of each component averaged over 1623 fragment pairs of 3DMatch in milli-seconds.}
	\label{res:comlexsity}
	\centering
	 \resizebox{\linewidth}{!}{
	% \small
	\begin{tabular}{c|cccccc}
		\toprule
		Method&data loader&encoder&attention layer&decoder&PGMO&FGMO\\
		\midrule
        Predator\cite{huang2021predator}&191& 9& 70&1& \XSolidBrush&\XSolidBrush\\
        GeoTransformer\cite{qin2022geometric}&-& -& 60&1& \XSolidBrush&\XSolidBrush\\
        GMONet&-&-&60&-&150&90\\
		\bottomrule
	\end{tabular}}
\vspace{-0.5cm}
\end{table}

% \vspace{-0.1cm}
\subsection{Outdoor dataset: KITTI Odometry}
\textbf{Dataset.}
KITTI Odometry benchmark\cite{geiger2013vision} consists of 11 sequences of point clouds scanned by LiDAR. We follow\cite{huang2021predator,qin2022geometric} to use sequences 0-5 for training, 6-7 for validation, and 8-10 for testing.

\textbf{Metrics.}
Following \cite{huang2021predator}, we evaluate GMONet with three metrics: (1) rigid Registration Recall (RR), the fraction of successful registration pairs ($i.e.,$ RRE$<5^\circ$ and RTE$<$2m). The definitions of Relative Rotation Error (RRE) and Relative Translation Error (RTE) are the same as used in the 3DMatch benchmark.

\textbf{Results.}
In the Tab.\ref{res:KITTI}, we compared GMONet with several state-of-the-art RANSAC-based methods: 3DFeat-Net\cite{yew20183dfeat}, FCGF\cite{choy2019fully}, D3Feat\cite{bai2020d3feat}, Predator\cite{huang2021predator}, CoFiNet\cite{yu2021cofinet}, and GeoTransformer\cite{qin2022geometric}. The quantitative results show that our method can handle outdoor scene registration and achieve competitive performance.

% \vspace{-0.1cm}
\section{Conclusion}
We proposed a novel framework integrating rigid isometry-preserving constraints in the point feature learning stage. Specifically, we used the partial graph matching constraint at the coarse level and the mini-batch full graph matching constraint at the fine level. Experimental results show that our method has competitive performance for point cloud registration tasks. In the future, we would like to verify our method on 2D-2D and 2D-3D tasks.

% \section*{ACKNOWLEDGMENT}
% We want to thank the reviewers for their helpful comments.

%%%%%%%%%%%%%%%%%%%%%%%%%%%%%%%%%%%%%%%%%%%%%%%%%%%%%%%%%%%%%%%%%%%%%%%%%%%%%%%%

{\normalem
\bibliographystyle{IEEEtran} 
\bibliography{root}

\begin{thebibliography}{10}
\providecommand{\url}[1]{#1}
\csname url@samestyle\endcsname
\providecommand{\newblock}{\relax}
\providecommand{\bibinfo}[2]{#2}
\providecommand{\BIBentrySTDinterwordspacing}{\spaceskip=0pt\relax}
\providecommand{\BIBentryALTinterwordstretchfactor}{4}
\providecommand{\BIBentryALTinterwordspacing}{\spaceskip=\fontdimen2\font plus
\BIBentryALTinterwordstretchfactor\fontdimen3\font minus
  \fontdimen4\font\relax}
\providecommand{\BIBforeignlanguage}[2]{{%
\expandafter\ifx\csname l@#1\endcsname\relax
\typeout{** WARNING: IEEEtran.bst: No hyphenation pattern has been}%
\typeout{** loaded for the language `#1'. Using the pattern for}%
\typeout{** the default language instead.}%
\else
\language=\csname l@#1\endcsname
\fi
#2}}
\providecommand{\BIBdecl}{\relax}
\BIBdecl

\bibitem{schonberger2016structure}
J.~L. Schonberger and J.-M. Frahm, ``Structure-from-motion revisited,'' in
  \emph{Proceedings of the IEEE conference on computer vision and pattern
  recognition}, 2016, pp. 4104--4113.

\bibitem{choi2015robust}
S.~Choi, Q.-Y. Zhou, and V.~Koltun, ``Robust reconstruction of indoor scenes,''
  in \emph{Proceedings of the IEEE Conference on Computer Vision and Pattern
  Recognition}, 2015, pp. 5556--5565.

\bibitem{zhang2015visual}
J.~Zhang and S.~Singh, ``Visual-lidar odometry and mapping: Low-drift, robust,
  and fast,'' in \emph{2015 IEEE International Conference on Robotics and
  Automation (ICRA)}.\hskip 1em plus 0.5em minus 0.4em\relax IEEE, 2015, pp.
  2174--2181.

\bibitem{drost2010model}
B.~Drost, M.~Ulrich, N.~Navab, and S.~Ilic, ``Model globally, match locally:
  Efficient and robust 3d object recognition,'' in \emph{2010 IEEE computer
  society conference on computer vision and pattern recognition}.\hskip 1em
  plus 0.5em minus 0.4em\relax Ieee, 2010, pp. 998--1005.

\bibitem{zeng2017multi}
A.~Zeng, K.-T. Yu, S.~Song, D.~Suo, E.~Walker, A.~Rodriguez, and J.~Xiao,
  ``Multi-view self-supervised deep learning for 6d pose estimation in the
  amazon picking challenge,'' in \emph{2017 IEEE international conference on
  robotics and automation (ICRA)}.\hskip 1em plus 0.5em minus 0.4em\relax IEEE,
  2017, pp. 1386--1383.

\bibitem{yang2020teaser}
H.~Yang, J.~Shi, and L.~Carlone, ``Teaser: Fast and certifiable point cloud
  registration,'' \emph{IEEE Transactions on Robotics}, vol.~37, no.~2, pp.
  314--333, 2020.

\bibitem{li2019net}
Q.~Li, S.~Chen, C.~Wang, X.~Li, C.~Wen, M.~Cheng, and J.~Li, ``Lo-net: Deep
  real-time lidar odometry,'' in \emph{Proceedings of the IEEE/CVF Conference
  on Computer Vision and Pattern Recognition}, 2019, pp. 8473--8482.

\bibitem{zhang2014loam}
J.~Zhang and S.~Singh, ``Loam: Lidar odometry and mapping in real-time.'' in
  \emph{Robotics: Science and Systems}, vol.~2, no.~9.\hskip 1em plus 0.5em
  minus 0.4em\relax Berkeley, CA, 2014, pp. 1--9.

\bibitem{lee2021deep}
J.~Lee, S.~Kim, M.~Cho, and J.~Park, ``Deep hough voting for robust global
  registration,'' in \emph{Proceedings of the IEEE/CVF International Conference
  on Computer Vision}, 2021, pp. 15\,994--16\,003.

\bibitem{bai2021pointdsc}
X.~Bai, Z.~Luo, L.~Zhou, H.~Chen, L.~Li, Z.~Hu, H.~Fu, and C.-L. Tai,
  ``Pointdsc: Robust point cloud registration using deep spatial consistency,''
  in \emph{Proceedings of the IEEE/CVF Conference on Computer Vision and
  Pattern Recognition}, 2021, pp. 15\,859--15\,869.

\bibitem{mei2021cotreg}
G.~Mei, X.~Huang, L.~Yu, J.~Zhang, and M.~Bennamoun, ``Cotreg: Coupled optimal
  transport based point cloud registration,'' \emph{arXiv preprint
  arXiv:2112.14381}, 2021.

\bibitem{shen2022reliable}
Y.~Shen, L.~Hui, H.~Jiang, J.~Xie, and J.~Yang, ``Reliable inlier evaluation
  for unsupervised point cloud registration,'' \emph{arXiv preprint
  arXiv:2202.11292}, 2022.

\bibitem{jiang2021sampling}
H.~Jiang, Y.~Shen, J.~Xie, J.~Li, J.~Qian, and J.~Yang, ``Sampling network
  guided cross-entropy method for unsupervised point cloud registration,'' in
  \emph{Proceedings of the IEEE/CVF International Conference on Computer
  Vision}, 2021, pp. 6128--6137.

\bibitem{choy2019fully}
C.~Choy, J.~Park, and V.~Koltun, ``Fully convolutional geometric features,'' in
  \emph{Proceedings of the IEEE/CVF International Conference on Computer
  Vision}, 2019, pp. 8958--8966.

\bibitem{bai2020d3feat}
X.~Bai, Z.~Luo, L.~Zhou, H.~Fu, L.~Quan, and C.-L. Tai, ``D3feat: Joint
  learning of dense detection and description of 3d local features,'' in
  \emph{Proceedings of the IEEE/CVF conference on computer vision and pattern
  recognition}, 2020, pp. 6359--6367.

\bibitem{thomas2019kpconv}
H.~Thomas, C.~R. Qi, J.-E. Deschaud, B.~Marcotegui, F.~Goulette, and L.~J.
  Guibas, ``Kpconv: Flexible and deformable convolution for point clouds,'' in
  \emph{Proceedings of the IEEE/CVF international conference on computer
  vision}, 2019, pp. 6411--6420.

\bibitem{huang2021predator}
S.~Huang, Z.~Gojcic, M.~Usvyatsov, A.~Wieser, and K.~Schindler, ``Predator:
  Registration of 3d point clouds with low overlap,'' in \emph{Proceedings of
  the IEEE/CVF Conference on computer vision and pattern recognition}, 2021,
  pp. 4267--4276.

\bibitem{yu2021cofinet}
H.~Yu, F.~Li, M.~Saleh, B.~Busam, and S.~Ilic, ``Cofinet: Reliable
  coarse-to-fine correspondences for robust pointcloud registration,''
  \emph{Advances in Neural Information Processing Systems}, vol.~34, pp.
  23\,872--23\,884, 2021.

\bibitem{qin2022geometric}
Z.~Qin, H.~Yu, C.~Wang, Y.~Guo, Y.~Peng, and K.~Xu, ``Geometric transformer for
  fast and robust point cloud registration,'' in \emph{Proceedings of the
  IEEE/CVF Conference on Computer Vision and Pattern Recognition}, 2022, pp.
  11\,143--11\,152.

\bibitem{li2022lepard}
Y.~Li and T.~Harada, ``Lepard: Learning partial point cloud matching in rigid
  and deformable scenes,'' in \emph{Proceedings of the IEEE/CVF Conference on
  Computer Vision and Pattern Recognition}, 2022, pp. 5554--5564.

\bibitem{wang2021you}
H.~Wang, Y.~Liu, Z.~Dong, W.~Wang, and B.~Yang, ``You only hypothesize once:
  Point cloud registration with rotation-equivariant descriptors,'' \emph{arXiv
  preprint arXiv:2109.00182}, 2021.

\bibitem{yew2022regtr}
Z.~J. Yew and G.~H. Lee, ``Regtr: End-to-end point cloud correspondences with
  transformers,'' in \emph{Proceedings of the IEEE/CVF Conference on Computer
  Vision and Pattern Recognition}, 2022, pp. 6677--6686.

\bibitem{vaswani2017attention}
A.~Vaswani, N.~Shazeer, N.~Parmar, J.~Uszkoreit, L.~Jones, A.~N. Gomez,
  {\L}.~Kaiser, and I.~Polosukhin, ``Attention is all you need,''
  \emph{Advances in neural information processing systems}, vol.~30, 2017.

\bibitem{zhou2015factorized}
F.~Zhou and F.~De~la Torre, ``Factorized graph matching,'' \emph{IEEE
  transactions on pattern analysis and machine intelligence}, vol.~38, no.~9,
  pp. 1774--1789, 2015.

\bibitem{gao2021deep}
Q.~Gao, F.~Wang, N.~Xue, J.-G. Yu, and G.-S. Xia, ``Deep graph matching under
  quadratic constraint,'' in \emph{Proceedings of the IEEE/CVF Conference on
  Computer Vision and Pattern Recognition}, 2021, pp. 5069--5078.

\bibitem{chen2022utopic}
Z.~Chen, H.~Chen, L.~Gong, X.~Yan, J.~Wang, Y.~Guo, J.~Qin, and M.~Wei,
  ``Utopic: Uncertainty-aware overlap prediction network for partial point
  cloud registration,'' \emph{arXiv preprint arXiv:2208.02712}, 2022.

\bibitem{peyre2016gromov}
G.~Peyr{\'e}, M.~Cuturi, and J.~Solomon, ``Gromov-wasserstein averaging of
  kernel and distance matrices,'' in \emph{International Conference on Machine
  Learning}.\hskip 1em plus 0.5em minus 0.4em\relax PMLR, 2016, pp. 2664--2672.

\bibitem{xu2019gromov}
H.~Xu, D.~Luo, H.~Zha, and L.~C. Duke, ``Gromov-wasserstein learning for graph
  matching and node embedding,'' in \emph{International conference on machine
  learning}.\hskip 1em plus 0.5em minus 0.4em\relax PMLR, 2019, pp. 6932--6941.

\bibitem{liu2020partial}
W.~Liu, C.~Zhang, J.~Xie, Z.~Shen, H.~Qian, and N.~Zheng, ``Partial
  gromov-wasserstein learning for partial graph matching,'' \emph{arXiv
  preprint arXiv:2012.01252}, 2020.

\bibitem{xie2020fast}
Y.~Xie, X.~Wang, R.~Wang, and H.~Zha, ``A fast proximal point method for
  computing exact wasserstein distance,'' in \emph{Uncertainty in artificial
  intelligence}.\hskip 1em plus 0.5em minus 0.4em\relax PMLR, 2020, pp.
  433--453.

\bibitem{nguyen2022improving}
K.~Nguyen, D.~Nguyen, T.~Pham, N.~Ho \emph{et~al.}, ``Improving mini-batch
  optimal transport via partial transportation,'' in \emph{International
  Conference on Machine Learning}.\hskip 1em plus 0.5em minus 0.4em\relax PMLR,
  2022, pp. 16\,656--16\,690.

\bibitem{besl1992method}
P.~J. Besl and N.~D. McKay, ``Method for registration of 3-d shapes,'' in
  \emph{Sensor fusion IV: control paradigms and data structures}, vol.
  1611.\hskip 1em plus 0.5em minus 0.4em\relax Spie, 1992, pp. 586--606.

\bibitem{yang2013go}
J.~Yang, H.~Li, and Y.~Jia, ``Go-icp: Solving 3d registration efficiently and
  globally optimally,'' in \emph{Proceedings of the IEEE International
  Conference on Computer Vision}, 2013, pp. 1457--1464.

\bibitem{zhou2016fast}
Q.-Y. Zhou, J.~Park, and V.~Koltun, ``Fast global registration,'' in
  \emph{European conference on computer vision}.\hskip 1em plus 0.5em minus
  0.4em\relax Springer, 2016, pp. 766--782.

\bibitem{zeng20173dmatch}
A.~Zeng, S.~Song, M.~Nie{\ss}ner, M.~Fisher, J.~Xiao, and T.~Funkhouser,
  ``3dmatch: Learning local geometric descriptors from rgb-d reconstructions,''
  in \emph{Proceedings of the IEEE conference on computer vision and pattern
  recognition}, 2017, pp. 1802--1811.

\bibitem{deng2018ppfnet}
H.~\vspace{0mm}Deng, T.~Birdal, and S.~Ilic, ``Ppfnet: Global context aware
  local features for robust 3d point matching,'' in \emph{Proceedings of the
  IEEE conference on computer vision and pattern recognition}, 2018, pp.
  195--205.

\bibitem{deng2018ppf}
H.~Deng, T.~Birdal, and S.~Ilic, ``Ppf-foldnet: Unsupervised learning of
  rotation invariant 3d local descriptors,'' in \emph{Proceedings of the
  European Conference on Computer Vision (ECCV)}, 2018, pp. 602--618.

\bibitem{zhao20193d}
Y.~Zhao, T.~Birdal, H.~Deng, and F.~Tombari, ``3d point capsule networks,'' in
  \emph{Proceedings of the IEEE/CVF Conference on Computer Vision and Pattern
  Recognition}, 2019, pp. 1009--1018.

\bibitem{cuturi2013sinkhorn}
M.~Cuturi, ``Sinkhorn distances: Lightspeed computation of optimal transport,''
  \emph{Advances in neural information processing systems}, vol.~26, 2013.

\bibitem{fu2021robust}
K.~Fu, S.~Liu, X.~Luo, and M.~Wang, ``Robust point cloud registration framework
  based on deep graph matching,'' in \emph{Proceedings of the IEEE/CVF
  Conference on Computer Vision and Pattern Recognition}, 2021, pp. 8893--8902.

\bibitem{kuhn1955hungarian}
H.~W. Kuhn, ``The hungarian method for the assignment problem,'' \emph{Naval
  research logistics quarterly}, vol.~2, no. 1-2, pp. 83--97, 1955.

\bibitem{wang2021neural}
R.~Wang, J.~Yan, and X.~Yang, ``Neural graph matching network: Learning
  lawler’s quadratic assignment problem with extension to hypergraph and
  multiple-graph matching,'' \emph{IEEE Transactions on Pattern Analysis and
  Machine Intelligence}, 2021.

\bibitem{zass2008probabilistic}
R.~Zass and A.~Shashua, ``Probabilistic graph and hypergraph matching,'' in
  \emph{2008 IEEE Conference on Computer Vision and Pattern Recognition}.\hskip
  1em plus 0.5em minus 0.4em\relax IEEE, 2008, pp. 1--8.

\bibitem{vayer2018fused}
T.~Vayer, L.~Chapel, R.~Flamary, R.~Tavenard, and N.~Courty, ``Fused
  gromov-wasserstein distance for structured objects: theoretical foundations
  and mathematical properties,'' \emph{arXiv preprint arXiv:1811.02834}, 2018.

\bibitem{titouan2019optimal}
V.~Titouan, N.~Courty, R.~Tavenard, and R.~Flamary, ``Optimal transport for
  structured data with application on graphs,'' in \emph{International
  Conference on Machine Learning}.\hskip 1em plus 0.5em minus 0.4em\relax PMLR,
  2019, pp. 6275--6284.

\bibitem{grave2019unsupervised}
E.~Grave, A.~Joulin, and Q.~Berthet, ``Unsupervised alignment of embeddings
  with wasserstein procrustes,'' in \emph{The 22nd International Conference on
  Artificial Intelligence and Statistics}.\hskip 1em plus 0.5em minus
  0.4em\relax PMLR, 2019, pp. 1880--1890.

\bibitem{alvarez2019towards}
D.~Alvarez-Melis, S.~Jegelka, and T.~S. Jaakkola, ``Towards optimal transport
  with global invariances,'' in \emph{The 22nd International Conference on
  Artificial Intelligence and Statistics}.\hskip 1em plus 0.5em minus
  0.4em\relax PMLR, 2019, pp. 1870--1879.

\bibitem{shen2021accurate}
Z.~Shen, J.~Feydy, P.~Liu, A.~H. Curiale, R.~San Jose~Estepar, R.~San
  Jose~Estepar, and M.~Niethammer, ``Accurate point cloud registration with
  robust optimal transport,'' \emph{Advances in Neural Information Processing
  Systems}, vol.~34, pp. 5373--5389, 2021.

\bibitem{eisenberger2020deep}
M.~Eisenberger, A.~Toker, L.~Leal-Taix{\'e}, and D.~Cremers, ``Deep shells:
  Unsupervised shape correspondence with optimal transport,'' \emph{Advances in
  Neural information processing systems}, vol.~33, pp. 10\,491--10\,502, 2020.

\bibitem{mandad2017variance}
M.~Mandad, D.~Cohen-Steiner, L.~Kobbelt, P.~Alliez, and M.~Desbrun,
  ``Variance-minimizing transport plans for inter-surface mapping,'' \emph{ACM
  Transactions on Graphics (ToG)}, vol.~36, no.~4, pp. 1--14, 2017.

\bibitem{frank1956algorithm}
M.~Frank and P.~Wolfe, ``An algorithm for quadratic programming,'' \emph{Naval
  research logistics quarterly}, vol.~3, no. 1-2, pp. 95--110, 1956.

\bibitem{zaslavskiy2008path}
M.~Zaslavskiy, F.~Bach, and J.-P. Vert, ``A path following algorithm for the
  graph matching problem,'' \emph{IEEE Transactions on Pattern Analysis and
  Machine Intelligence}, vol.~31, no.~12, pp. 2227--2242, 2008.

\bibitem{fatras2019learning}
K.~Fatras, Y.~Zine, R.~Flamary, R.~Gribonval, and N.~Courty, ``Learning with
  minibatch wasserstein: asymptotic and gradient properties,'' \emph{arXiv
  preprint arXiv:1910.04091}, 2019.

\bibitem{fatras2021unbalanced}
K.~Fatras, T.~S{\'e}journ{\'e}, R.~Flamary, and N.~Courty, ``Unbalanced
  minibatch optimal transport; applications to domain adaptation,'' in
  \emph{International Conference on Machine Learning}.\hskip 1em plus 0.5em
  minus 0.4em\relax PMLR, 2021, pp. 3186--3197.

\bibitem{benamou2015iterative}
J.-D. Benamou, G.~Carlier, M.~Cuturi, L.~Nenna, and G.~Peyr{\'e}, ``Iterative
  bregman projections for regularized transportation problems,'' \emph{SIAM
  Journal on Scientific Computing}, vol.~37, no.~2, pp. A1111--A1138, 2015.

\bibitem{dykstra1983algorithm}
R.~L. Dykstra, ``An algorithm for restricted least squares regression,''
  \emph{Journal of the American Statistical Association}, vol.~78, no. 384, pp.
  837--842, 1983.

\bibitem{geiger2013vision}
A.~Geiger, P.~Lenz, C.~Stiller, and R.~Urtasun, ``Vision meets robotics: The
  kitti dataset,'' \emph{The International Journal of Robotics Research},
  vol.~32, no.~11, pp. 1231--1237, 2013.

\bibitem{paszke2019pytorch}
A.~Paszke, S.~Gross, F.~Massa, A.~Lerer, J.~Bradbury, G.~Chanan, T.~Killeen,
  Z.~Lin, N.~Gimelshein, L.~Antiga \emph{et~al.}, ``Pytorch: An imperative
  style, high-performance deep learning library,'' \emph{Advances in neural
  information processing systems}, vol.~32, 2019.

\bibitem{yew20183dfeat}
Z.~J. Yew and G.~H. Lee, ``3dfeat-net: Weakly supervised local 3d features for
  point cloud registration,'' in \emph{Proceedings of the European conference
  on computer vision (ECCV)}, 2018, pp. 607--623.

\end{thebibliography}
}

\end{document}